\documentclass[10pt,twocolumn,letterpaper]{article}

\usepackage{cvpr}              

\usepackage{multirow}

\definecolor{cvprblue}{rgb}{0.21,0.49,0.74}
\usepackage[pagebackref,breaklinks,colorlinks,allcolors=cvprblue]{hyperref}

\def\paperID{} 
\def\confName{CVPR}
\def\confYear{2026}

\title{Aesthetic Camera Viewpoint Suggestion with 3D Aesthetic Field}

\author{Sheyang Tang\textsuperscript{1}, Armin Shafiee Sarvestani\textsuperscript{1}, Jialu Xu\textsuperscript{1}, Xiaoyu Xu\textsuperscript{2}, Zhou Wang\textsuperscript{1}\\
\textsuperscript{1}University of Waterloo, \textsuperscript{2}City University of Hong Kong\\
{\tt\small \{sheyang.tang,a5shafie,jialu.xu,zhou.wang\}@uwaterloo.ca}
}

\begin{document}

\twocolumn[{%
\renewcommand\twocolumn[1][]{#1}%
\maketitle
\centering
\includegraphics[width=0.93\linewidth]{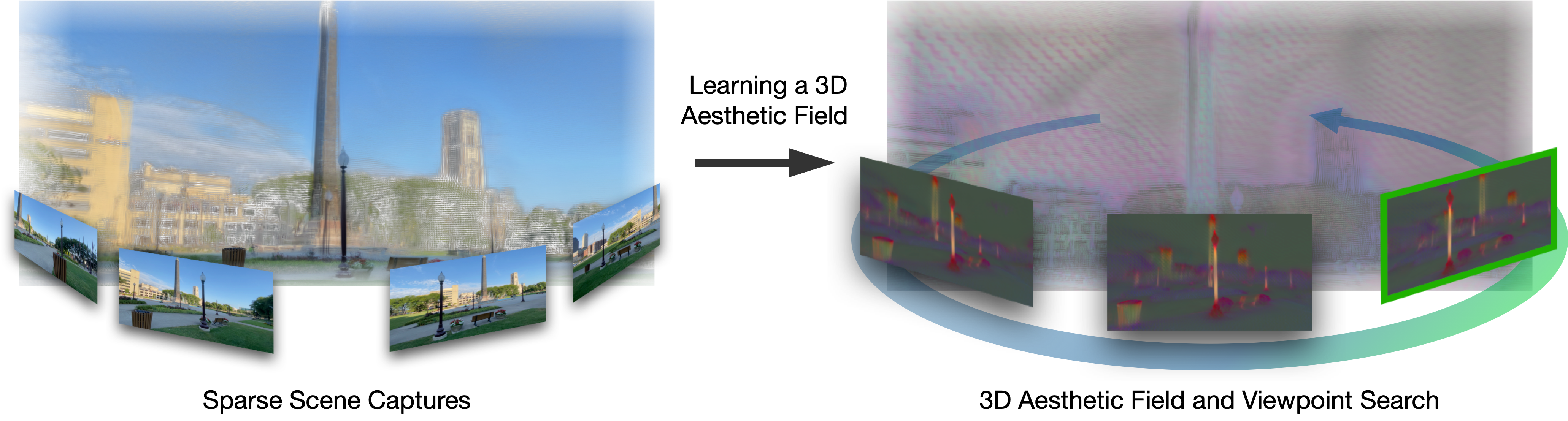}
\captionof{figure}{Given sparse scene captures (\textit{left}), our method learns a 3D aesthetic field that encodes spatially varying aesthetic cues.
This field enables geometry-grounded aesthetic reasoning in 3D, allowing efficient discovery of appealing camera viewpoints (\textit{right}). \vspace{1em}}
\label{fig:teaser}
}]


\begin{abstract}
The aesthetic quality of a scene depends strongly on camera viewpoint.
Existing approaches for aesthetic viewpoint suggestion are either single-view adjustments, predicting limited camera adjustments from a single image without understanding scene geometry, or 3D exploration approaches, which rely on dense captures or prebuilt 3D environments coupled with costly reinforcement learning (RL) searches. In this work, we introduce the notion of 3D aesthetic field that enables geometry-grounded aesthetic reasoning in 3D with sparse captures, allowing efficient viewpoint suggestions in contrast to costly RL searches.
We opt to learn this 3D aesthetic field using a feedforward 3D Gaussian Splatting network that distills high-level aesthetic knowledge from a pretrained 2D aesthetic model into 3D space, enabling aesthetic prediction for novel viewpoints from only sparse input views. Building on this field, we propose a two-stage search pipeline that combines coarse viewpoint sampling with gradient-based refinement, efficiently identifying aesthetically appealing viewpoints without dense captures or RL exploration. Extensive experiments show that our method consistently suggests viewpoints with superior framing and composition compared to existing approaches, establishing a new direction toward 3D-aware aesthetic modeling.
\end{abstract}

\section{Introduction}
\label{sec:intro}

Framing and composition play a central role in image aesthetics.
The same 3D scene can appear engaging or awkward depending solely on the camera’s viewpoint, as spatial relationships and perspective vary with viewing position. This makes aesthetics inherently \textit{3D-dependent}.  
When framing a shot, experienced photographers intuitively reason about spatial layout by observing from multiple angles and forming an internal sense of scene aesthetics to anticipate how visual appeal changes with viewpoint~\cite{freeman2017photographer}. This can be viewed as developing a mental map of aesthetic variation across viewpoints.
Enabling machines to reason in a similar way—inferring spatially-varying scene aesthetics from a few observations to suggest appealing viewpoints—would benefit not only personal photography~\cite{su2021camera,li2025towards} but also view selection/planning in VR/AR~\cite{wang2025cinefolio,zaman2023vicarious} and autonomous systems~\cite{xie2023gait,alzayer2021autophoto}.

Existing approaches primarily focus on \textit{single-view adjustments}~\cite{li2025towards,su2021camera,liu2023beyond} by predicting limited camera movements to refine a single image. Although they can enhance framing locally, they lack awareness of the underlying scene geometry, leaving their reasoning confined to a narrow neighborhood around the anchor view.
Recent works attempt to incorporate broader viewpoint changes into aesthetic modeling~\cite{yao2025photography,uchida20253d} using generative models. However, they rely on hallucinated content from a single view and thus cannot ensure geometric consistency with real scenes. This highlights the need for a 3D-aware solution grounded in scene geometry for reliable aesthetic reasoning beyond observed views in 3D space.

In another line of work, \textit{3D exploration} approaches explicitly operate in spatial environments, employing reinforcement-learning (RL) or genetic algorithms to search for appealing  viewpoints~\cite{alzayer2021autophoto,xie2023gait,wu2024viewactive,skartados2024finding}. However, they assume access to dense, high-quality visual inputs from either real or readily available virtual 3D environments (\eg simulations, pretrained NeRFs~\cite{mildenhall2021nerf}), which are costly to construct and require extensive data collection. Moreover, their RL-based solutions involve costly step-by-step exploration in the environment. These limitations call for a solution that alleviates the dependency on dense captures and enables efficient aesthetic reasoning without iterative physical adjustments in real world.

To address these limitations, we propose to unify aesthetic perception with 3D geometry understanding by learning a \textit{3D aesthetic field} that enables efficient inference of scene aesthetics from sparse observations.
This field intrinsically encodes spatially varying aesthetic cues, supporting geometry-grounded aesthetic reasoning in 3D.
Much like a skilled photographer who observes the scene from a few angles and develops a mental map of where the best angles lie, our system learns a 3D aesthetic field from sparse observations to reason about aesthetic variation across viewpoints and find appealing camera poses, as shown in~\cref{fig:teaser}.
Specifically, we distill knowledge from a pretrained 2D aesthetic model~\cite{wei2018good} into a feedforward 3D Gaussian Splatting network~\cite{depthsplat}, predicting per-Gaussian aesthetic features directly from sparse input views. Rendering aesthetic features from this field allows evaluation of aesthetic quality at novel viewpoints. The learned aesthetic field transforms viewpoint suggestion into a differentiable optimization problem, which we address by an efficient two-stage search pipeline: (1) we sample candidate viewpoints, and (2) locally refine them through gradient-based updates. By unifying aesthetic perception with geometric understanding, our method reasons beyond the observed views to efficiently identify optimal viewpoints within the scene, without the overhead of RL or dense captures for reconstruction. 
Extensive experiments demonstrate the superiority of our framework in aesthetic viewpoint suggestion.
Our main contributions are:
\begin{itemize}
    \item We introduce the task of \textit{3D-aware} aesthetic viewpoint suggestion with \textit{sparse observation}s, accounting for \textit{3D-dependency} in aesthetic modeling without requiring \textit{dense captures}.

    \item We propose a novel \textit{3D aesthetic field} that unifies 2D aesthetic perception with 3D geometric understanding, modeling aesthetic variation across viewpoints.

    \item We develop an \textit{efficient} two-stage search pipeline that combines coarse sampling with gradient-based refinement to efficiently discover appealing viewpoints.

    \item Extensive experiments on diverse datasets and input settings demonstrate the effectiveness of our framework for 3D-aware aesthetic viewpoint suggestion.
\end{itemize}




\section{Related Works}
\label{sec:related}
\subsection{Single-view Adjustment Methods}
Early aesthetic optimization works focus on image cropping~\cite{wei2018good,zeng2019reliable,hong2021composing,wang2023image} to reframe an existing photo to improve visual appeal. They predict cropping windows to maximize the aesthetic score but function only as post-processing once an image is captured. Later works extend aesthetic adjustment to virtual camera motion, reusing cropping datasets for learning camera adjustment that emulate crop movements. Su~\etal~\cite{su2021camera} predict in-plane camera shifts for interactive composition refinement, while Li~\etal~\cite{li2025towards} suggest camera rotations through an iterative feedback loop. Liu~\etal~\cite{liu2023beyond} employ outpainting to suggest cropping windows beyond the image border, but their model assumes a fixed camera with enlarged field of view, lacking true parallax changes. More recent efforts attempt to incorporate broader viewpoint variation. Uchida~\etal~\cite{uchida20253d} combine outpainting with single-image 3D reconstruction to enlarge the search space, and Yao~\etal~\cite{yao2025photography} synthesize poor-to-better perspectives via image-to-video generation. However, they rely on hallucinated content and cannot ensure geometric consistency with real scenes.
Lacking awareness of the underlying 3D structure, single-view adjustment methods are restricted to local composition refinement around the original viewpoint. They struggle to discover more aesthetic angles that involve excluding or introducing scene elements, which requires 3D reasoning beyond observed views. In contrast, our method explicitly models spatially varying scene aesthetics, supporting geometry-aware aesthetic reasoning in 3D.

\subsection{3D-exploration Methods}
3D-exploration methods reason directly in 3D environments. They employ RL or search algorithms to actively explore candidate viewpoints in real or simulated scenes. AutoPhoto~\cite{alzayer2021autophoto} uses RL to navigate through 3D to find aesthetic views, while GAIT~\cite{xie2023gait} trains an RL agent to generate indoor tours in simulation. ViewActive~\cite{wu2024viewactive} uses geometric and semantic cues to optimize viewpoint selection for mesh objects. Skartados~\etal~\cite{skartados2024finding} adopt a genetic algorithm to find aesthetic views in a pretrained NeRF scene. However, they rely on dense captures or prebuilt 3D assets that are costly to obtain, and their iterative RL procedures requires physical adjustments in real world. In contrast, our framework benefits from efficient inference of 3D scene aesthetics from sparse observations, and virtually reasons within the aesthetic field. 

\subsection{Feature Distillation in 3D Gaussian Splatting}
Recent advances in 3D Gaussian Splatting~\cite{kerbl20233d,mvsplat,depthsplat} have enabled efficient, differentiable rendering and provided a versatile representation for encoding scene geometry and appearance. Building on this, several studies have explored distilling 2D semantic features~\cite{qin2024langsplat,zhou2024feature,li2025semanticsplat,tian2025uniforward} into 3D Gaussian fields for segmentation tasks. These developments suggest the potential of transferring 2D knowledge into 3D representations. However, unlike semantic features that are largely viewpoint-invariant, aesthetic information is inherently viewpoint-dependent and remains largely unexplored in this context. Our work extends this paradigm by distilling aesthetic perception into 3D Gaussian fields to enable viewpoint-aware aesthetic modeling.

\begin{figure*}[htbp]
\centering
    \includegraphics[width=\linewidth]{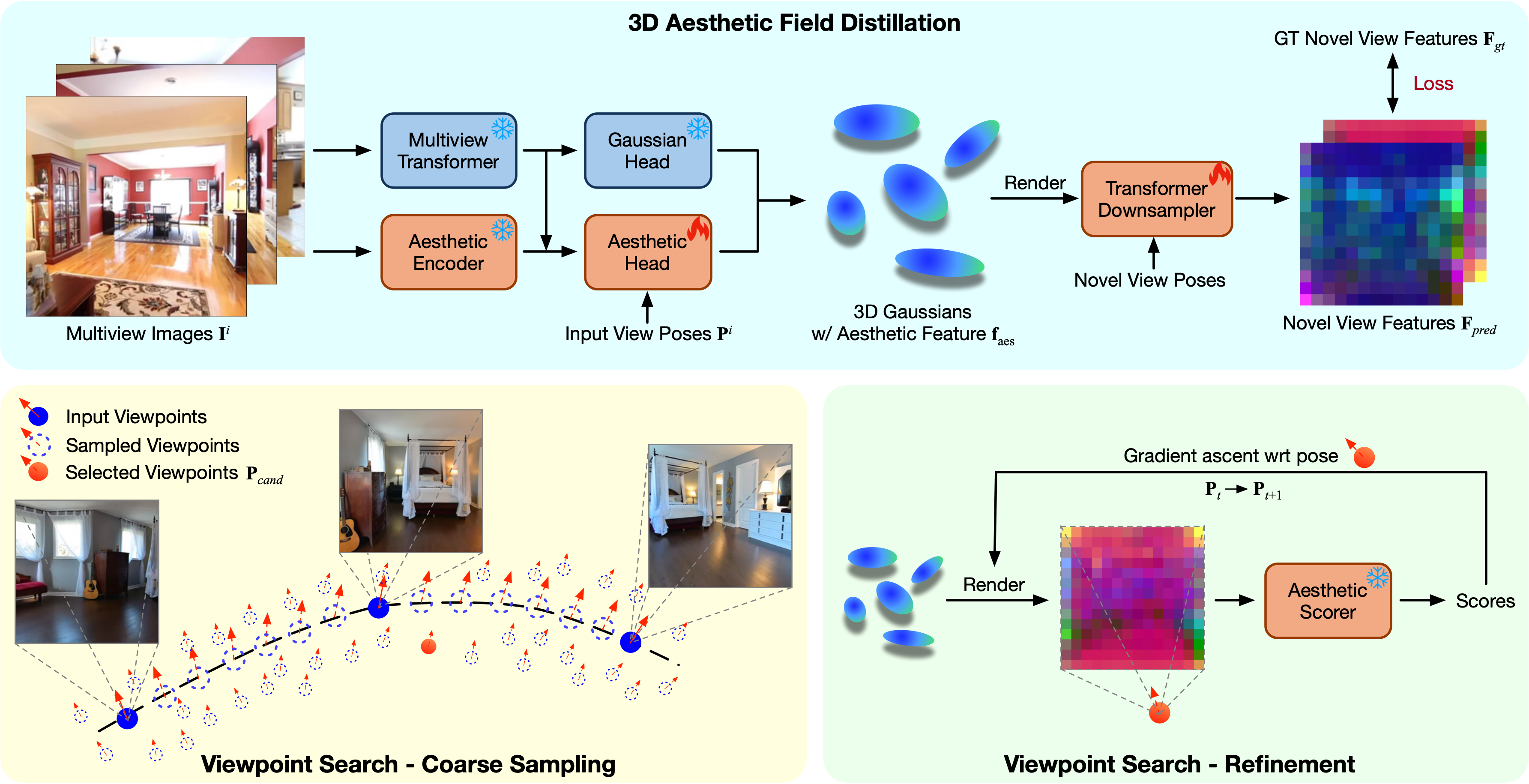}
    \caption{We distill aesthetic features into a feedforward Gaussian Splatting network (\textit{top}). At inference, to search for aesthetic viewpoints, we adopt a two-stage pipeline: coarse sampling to find good candidates (\textit{bottom left}) and local refinement by gradient ascent (\textit{bottom right}).}
    \label{fig:framework}
\end{figure*}

\section{Proposed Method}
\label{sec:method}

\textbf{Problem Statement.} We aim to identify aesthetically pleasing viewpoints for a given 3D scene observed from sparse input views. We formulate this as finding the camera pose $\mathbf P^*$ that maximizes the predicted aesthetic score $score(\mathbf P)$ over the continuous view space:
\begin{equation}
    \mathbf P^* = \arg\max_\mathbf P score(\mathbf P).
\end{equation}
Since direct optimization is intractable as it requires back-and-forth 2D projection and evaluation, we propose to approximate this objective by learning a \textit{3D aesthetic field}, which defines a continuous mapping from camera pose to aesthetic quality while being grounded in scene geometry.

To learn this field, we \textit{distill} aesthetic knowledge from a pretrained 2D aesthetic model into a feedforward Gaussian Splatting framework~\cite{depthsplat,mvsplat} to reconstruct per-Gaussian geometry and aesthetic features from sparse input views.
The distilled field enables rendering of aesthetic features at novel viewpoints, which are evaluated by an aesthetic decoder to assess framing and composition.
To efficiently explore the 3D scene for viewpoint suggestion, we introduce a two-stage search pipeline: (1) coarse sampling of candidate viewpoints along and around the input trajectory followed by aesthetic scoring to select top candidates, and (2) gradient-based optimization for local refinement.
The overall framework is illustrated in \cref{fig:framework}.

\subsection{Distilling the 3D Aesthetic Field}\label{sec:distill}
\textbf{Feedforward 3D Gaussian Splatting.}
We build our framework upon a feedforward Gaussian Spatting model~\cite{depthsplat} that predicts per-pixel Gaussian representation from sparse input views in a single forward pass. Specifically, given input views $\{\mathbf I^i\in\mathbb{R}^{H\times W\times 3}\}_{i=1}^N$ and corresponding camera poses $\{\mathbf P^i\in\mathbb{R}^{3\times 4}\}_{i=1}^N$, a multi-view transformer extracts multi-view features $\{\mathbf F^i_{mv}\}_{i=1}^N$. The features are then fused across neighboring views using plane-sweep~\cite{collins1996space,xu2023unifying} aggregation and enriched with monocular depth cues to predict per-pixel depth, which determines the center $\boldsymbol{\mu}$ of each 3D Gaussian. A DPT head~\cite{dpt} finally regresses the remaining per-Gaussian parameters including covariance $\boldsymbol\Sigma$, opacity $\alpha$, and color $\mathbf c$. Novel view images can be rendered from the 3D Gaussians via Gaussian Splatting~\cite{kerbl20233d}.
This backbone provides a differentiable and efficient 3D representation that we extend to build the aesthetic field.

\noindent\textbf{Direct RGB Scoring and Its Instability.}
Given the feedforward Gaussian Splatting backbone, a straightforward solution for viewpoint suggestion is to render novel views and directly score them using a pretrained aesthetic model.
However, this naive approach faces two key challenges.
\textit{First}, aesthetic models are highly sensitive to small pixel variations, producing fluctuating scores even for adjacent views (see \cref{fig:nva_1}). This is because existing datasets lack annotations across nearby viewpoints in 3D, hence the models haven't seen such variations in training.
\textit{Second}, rendering artifacts at novel views can bias aesthetic predictions and misguide optimization (see \cref{fig:nva_1}).
To address them, we shift from pixel-level scoring to feature-level reasoning by distilling aesthetic representations into the 3D Gaussians.
Operating in this latent feature space enhances robustness to low-level artifacts and enforces multi-view spatial consistency, leading to smoother aesthetic variations across nearby views. See \cref{sec:nva} for further details.

\noindent\textbf{Aesthetic Feature Distillation.}
We distill high-level aesthetic representations from a pretrained aesthetic teacher model~\cite{wei2018good}.
Specifically, we select an intermediate layer as the distillation target and train our model to predict a per-Gaussian aesthetic embedding $\mathbf f_{aes}$, which can be decoded into aesthetic scores through the remaining layers of the teacher after rasterization.

As shown in \cref{fig:framework}, on top of the backbone, we introduce lightweight aesthetic modules: a CNN aesthetic encoder, an aesthetic DPT head~\cite{dpt}, and a transformer downsampler. The aesthetic encoder is from the teacher's feature extraction layers~\cite{wei2018good} and produces multi-scale aesthetic features $\{\mathbf F_{aes}^i\}_{i=1}^N$ from the input images. These features are fused with multi-view features $\{\mathbf F^i_{mv}\}_{i=1}^N$ from the backbone, and processed by the aesthetic DPT head~\cite{dpt} to regress per-Gaussian aesthetic embeddings $\mathbf f_{aes}$. Combined with other Gaussian attributes $(\boldsymbol{\mu},\boldsymbol{\Sigma},\alpha)$, these embeddings are rendered into aesthetic feature maps $\mathbf{\widehat{F}}_{pred}$ at novel views through the same rasterization pipeline as RGB rendering. To reduce storage and rasterization overhead, each $\mathbf f_{aes}$ is a compact $32$-dim vector instead of $512$-dim as the teacher feature. Since the teacher feature maps $\mathbf{F}_{gt}$ are much smaller and deeper than $\mathbf{\widehat{F}}_{pred}$, we employ a lightweight transformer downsampler to align them and obtain the final prediction $\mathbf F_{pred}$. 

Furthermore, we condition the model on camera poses at both input and novel views, since aesthetic representations are inherently viewpoint-dependent. Incorporating pose information allows the model to capture these dependencies.

\noindent\textbf{Training Objective.}
The backbone multi-view transformer, DPT head, and aesthetic encoder are kept frozen to preserve consistent geometry predictions and 2D aesthetic perception, while the additional modules are trained end-to-end. We use mean-squared error (MSE) loss on the rendered feature maps $\mathbf F_{pred}$ at hold-out views, where ground-truth $\mathbf{F}_{gt}$ are obtained by feeding ground-truth images into the pretrained aesthetic model~\cite{wei2018good}.

\subsection{Viewpoint Search Pipeline}\label{sec:search}
The learned 3D aesthetic field provides a continuous and differentiable mapping from camera pose to aesthetic score, enabling efficient search and optimization in viewpoint space.
To efficiently explore this space, we adopt a two-stage coarse-to-fine search pipeline.
In the first stage, candidate viewpoints are heuristically sampled around the input trajectories and scored to identify high-quality candidates.
In the second stage, we perform gradient ascent on the selected candidates to locally refine their poses.
The distilled aesthetic field yields a smooth score landscape (see \cref{fig:nva_1} (a)), allowing stable gradient-based optimization (see \cref{sec:grad}).
The final refined viewpoints are returned as aesthetic suggestions. The entire process is illustrated in the bottom of~\cref{fig:framework}. We now describe each stage in detail.

\noindent\textbf{Stage 1: Coarse Sampling.}
The goal of this stage is to efficiently identify promising viewpoint candidates for further refinement.
We first connect the input views into a continuous camera trajectory by interpolating both camera positions and orientations, yielding a smooth path that covers the major observed viewpoints of the scene.
Candidate viewpoints are then linearly sampled along this trajectory.
For each sample, we further generate a local neighborhood of perturbed cameras with small in-plane shifts and directional jitters, allowing local exploration around the trajectory while maintaining scene focus.
Each candidate viewpoint is rendered through the aesthetic field to obtain its aesthetic features, which are evaluated by the aesthetic decoder to produce a score.
Finally, we select the top-$K$ high-scoring candidates $\{\mathbf{P}_{cand}^k\}_{k=1}^K$ for the subsequent refinement stage. To enhance diversity, near-identical candidates are filtered out via a distance-based duplication check, ensuring that the selected set covers distinct viewpoints.


\noindent\textbf{Stage 2: Gradient-based Refinement.} Starting from the candidate viewpoints obtained in Stage 1, we perform local optimization directly on camera poses to further maximize their aesthetic scores.
Specifically, each camera pose $\mathbf P_{cand}^k$ is parameterized by 3D translation and orientation, and updated through gradient ascent on the aesthetic score:
\begin{equation}
    \mathbf P_{t+1}=\mathbf P_{t} + \eta \nabla_{\mathbf P}score(\mathbf P_{t}),
\end{equation}
where $\eta$ is the step size.
In practice, we optimize a 5-dim vector comprising 3D translation and two rotation variables (yaw and pitch), as roll is rarely adjusted in typical scene captures. 
We use a fixed number of iterations for each candidate and retain the top-scoring refined viewpoints as the final suggestions.
\section{Experiments}
\label{sec:experiment}

\subsection{Experiment Setup}
\noindent\textbf{Datasets.}
We use RealEstate10k (RE10k)~\cite{re10k} and DL3DV~\cite{dl3dv} for training and evaluation. RE10k mainly contains indoor videos, while DL3DV involves more diverse scenes. Both have camera parameters at each frame. More details can be found in the Supplementary Material.

\noindent\textbf{Implementation Details.} We implement our framework in Pytorch~\cite{paszke2019pytorch}, using DepthSplat~\cite{depthsplat} as the feedforward Gaussian Splatting backbone. We use the VEN~\cite{wei2018good} model as the teacher for distilling aesthetic representations. VEN is a CNN model, and we use features from the $23^{rd}$ layer($14\times 14\times 512$) as the distillation target. Following Xu~\etal~\cite{depthsplat}, we use image resolutions of $256\times256$ for RE10k and $256\times448$ for DL3DV. During training, we randomly sample 2 input views for RE10k and 2 to 6 input views for DL3DV, following the same protocol as Xu~\etal~\cite{depthsplat}. More implementation and training details can be found in the Supplementary Material.


\noindent\textbf{Search Pipeline Configurations.} At inference, we search for optimal views via a two-stage approach. In Stage 1, within each segment of the interpolated trajectory, we uniformly sample 16 camera poses, and around each we further sample 8 neighboring poses. We take top-2 candidates as input to the next stage. In Stage 2, we use the Adam~\cite{kingma2014adam} optimizer with step size 0.01 and update for 25 steps.
\begin{table}[t]
\small
  \centering
  \begin{subtable}[t]{0.9\linewidth}
  \begin{tabular}{l| c | c c c c}
    \toprule
    \multirow{2}{*}{Methods} & \multirow{2}{*}{\#Views} & \multicolumn{2}{c}{RE10k~\cite{re10k}} & \multicolumn{2}{c}{DL3DV~\cite{dl3dv}} \\
     & & PLCC & SRCC & PLCC & SRCC \\
    \midrule
    Baseline & \multirow{2}{*}{2} & 0.657 & 0.628 & 0.326 & 0.307 \\
    \textbf{Ours} & & \textbf{0.780} & \textbf{0.740} &  \textbf{0.509} & \textbf{0.477} \\
    \midrule
    Baseline & \multirow{2}{*}{4} & 0.657 & 0.633 & 0.513 & 0.481\\
    \textbf{Ours} & & \textbf{0.796} & \textbf{0.758} &  \textbf{0.722} & \textbf{0.682} \\
    \midrule
    Baseline & \multirow{2}{*}{6} & 0.745 & 0.701 & 0.580 & 0.553\\
    \textbf{Ours} & & \textbf{0.836} & \textbf{0.794} &  \textbf{0.753} & \textbf{0.719}\\
    \bottomrule
  \end{tabular}
  \caption{Image resolution $256\times 256$.}
  \label{tab:correlation256}
  \end{subtable}
  \vfill
  \begin{subtable}[t]{0.9\linewidth}
  \begin{tabular}{l| c | c c c c}
    \toprule
    \multirow{2}{*}{Methods} & \multirow{2}{*}{\#Views} & \multicolumn{2}{c}{RE10k~\cite{re10k}} & \multicolumn{2}{c}{DL3DV~\cite{dl3dv}} \\
     & & PLCC & SRCC & PLCC & SRCC \\
    \midrule
    Baseline & \multirow{2}{*}{2} & 0.634 & 0.611 & 0.299 & 0.286 \\
    \textbf{Ours} & & \textbf{0.733} & \textbf{0.704} & \textbf{0.479} & \textbf{0.445} \\
    \midrule
    Baseline & \multirow{2}{*}{4} & 0.596 & 0.578 & 0.552 & 0.535 \\
    \textbf{Ours} & & \textbf{0.753} & \textbf{0.722} & \textbf{0.700} & \textbf{0.668} \\
    \midrule
    Baseline & \multirow{2}{*}{6} & 0.711 & 0.656 & 0.646 & 0.625 \\
    \textbf{Ours} & & \textbf{0.821} & \textbf{0.764} & \textbf{0.737} & \textbf{0.707} \\
    \bottomrule
  \end{tabular}
  \caption{Image resolution $256\times 448$.}
  \label{tab:correlation448}
\end{subtable}
\caption{Novel view aesthetic score correlation between predicted and ground-truth scores.}
\label{tab:correlation}
\end{table}

\begin{figure}[t]
\centering
    \includegraphics[width=\linewidth]{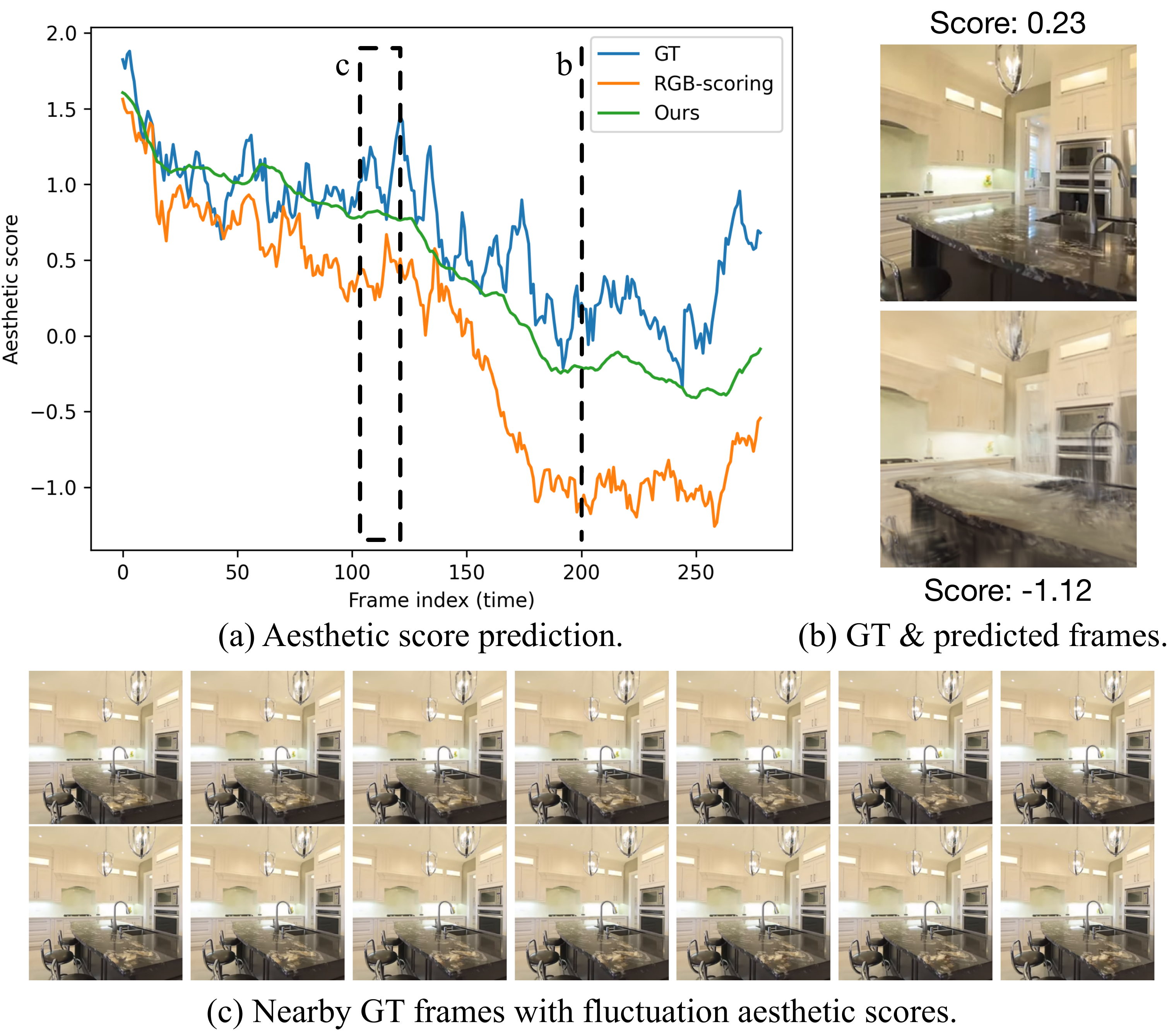}
    \caption{(a) Aesthetic score predictions over consecutive frames. Our method produces scores closer to the ground truth while being smoother and more consistent across nearby views. The dashed line and box mark the regions visualized in (b) and (c), respectively. (b) Given the same aesthetic model and viewpoint, rendering artifacts in the predicted view (\textit{bottom}) bias the RGB-scoring approach toward lower scores. (c) Ground truth scores fluctuate noticeably across nearly identical nearby views.}
    \label{fig:nva_1}
\end{figure}

\subsection{Aesthetic Prediction at Novel Views}\label{sec:nva}
We first evaluate our model’s ability to predict aesthetic quality at unseen viewpoints to validate the learned 3D aesthetic field before applying it to viewpoint suggestion.  
Specifically, we compare our predicted aesthetic scores with ground-truth scores from the teacher aesthetic model~\cite{wei2018good}, and benchmark against the RGB-scoring baseline.  
For each scene in the RE10k and DL3DV test splits, we use 2, 4, and 6 input views to build the aesthetic field and predict aesthetic scores for the remaining frames, using the entire video clips (about 250 frames each) for RE10k and 100 frames for DL3DV. Following Xu~\etal~\cite{depthsplat}, we train at $256\times 256$ in RE10k and $256\times 448$ in DL3DV, and evaluate under both resolutions to assess generalization.


\begin{table*}[htbp]
\small
  \centering
  \begin{tabular}{l| c c  | c c  | c c }
    \multicolumn{7}{c}{\textit{RE10k}~\cite{re10k}} \\
    \toprule
    \multirow{2}{*}{Methods} 
    & \multicolumn{2}{c|}{2 Input Views} & \multicolumn{2}{c|}{4 Input Views} & \multicolumn{2}{c}{6 Input Views} \\  
    & VEN$\uparrow$ & SAMPNet$\uparrow$ & VEN$\uparrow$ & SAMPNet$\uparrow$ & VEN$\uparrow$ & SAMPNet$\uparrow$ \\
    \midrule
    Baseline & 1.48 & 2.29  & 1.79 & 2.26  & 2.01 & 2.29  \\
    In-plane Shift\small $^{\ast}$& 1.52 & 2.31  & 1.81 & 2.36  & 2.10 & 2.41  \\
    Rotation\small $^{\ast}$ & 1.78 & 2.38  & 1.95 & 2.42  & 2.13 & 2.45  \\
    UNIC~\cite{liu2023beyond}\small $^{\dagger}$ &1.15 & 2.17  & 1.61 & 2.33  & 1.82 & 2.35  \\
    Uchida~\etal~\cite{uchida20253d}\small $^{\dagger}$ &1.58 & 2.32  & 1.89 & 2.37  & 2.13 & 2.42  \\
    \textbf{Ours} &\textbf{1.89} & \textbf{2.40}  & \textbf{2.03} & \textbf{2.45} & \textbf{2.20} & \textbf{2.49}  \\
    \bottomrule
    \multicolumn{7}{c}{\textit{DL3DV}~\cite{dl3dv}} \\
    \toprule
    Baseline &2.08 & 2.23 & 2.31 & 2.26  & 2.47 & 2.30  \\
    In-plane Shift\small $^{\ast}$ &2.16 & 2.28 & 2.45 & 2.29  & 2.59 & 2.33  \\
    Rotation\small $^{\ast}$ &2.52 & 2.30  & 2.67 & 2.32  & 2.85 & 2.37  \\
    UNIC~\cite{liu2023beyond}\small $^{\dagger}$ &2.07 & 2.21  & 2.35 & 2.27  & 2.48 & 2.30  \\
    Uchida~\etal~\cite{uchida20253d}\small $^{\dagger}$ &2.34 & 2.25  & 2.68 & 2.32 & 2.81 & 2.36  \\
    \textbf{Ours} &\textbf{2.56} & \textbf{2.33}  & \textbf{2.76} & \textbf{2.36}  & \textbf{2.91} & \textbf{2.41}  \\
    \bottomrule
  \end{tabular}
  \caption{Aesthetic viewpoint suggestion evaluation on different datasets and with different numbers of input views.  
    $^{\dagger}$ indicates open-sourced methods adapted to our setting.
    $^{\ast}$ denotes approximations of non–open-sourced single-view methods in our setting. Note that such approximations may overestimate the actual results that those methods could deliver. 
    }
  \label{tab:avs}
\end{table*}

\noindent\textbf{Correlation to Teacher Scores.} We compute the correlation between the predicted and teacher aesthetic scores at novel views using two standard metrics: the Pearson linear correlation coefficient (PLCC) and the Spearman rank-order correlation coefficient (SRCC), which are computed for each scene and then averaged. 
As shown in~\cref{tab:correlation}, across all settings, our aesthetic field achieves significantly higher correlation with the teacher than the RGB-scoring baseline, indicating that the distilled field produces more stable and faithful aesthetic predictions. 
As discussed earlier, direct RGB scoring is highly sensitive to rendering artifacts.  
Moreover, teacher scores themselves fluctuate across nearby ground-truth views with almost identical compositions, so these quantitative results are best interpreted as \textit{relative indicators} of stability and fidelity rather than absolute measures of accuracy.
To better illustrate these effects and the advantages of our feature-distilled field, we present qualitative comparisons below.

\noindent\textbf{Qualitative Comparisons.} \Cref{fig:nva_1} shows representative examples to illustrate the problems of RGB-scoring baseline and the superiority of our distilled aesthetic field.  
\textit{First}, rendering artifacts in predicted novel-view images (\eg noise, blurriness) can mislead the aesthetic model (\cref{fig:nva_1} (b)).
\textit{Second}, ground-truth views with nearly identical content often yield fluctuating teacher scores (\cref{fig:nva_1} (c)), revealing the inherent sensitivity of aesthetic models to pixel-level variations.
As shown in \cref{fig:nva_1} (a), these factors cause large oscillations in the RGB-scoring predictions (orange), whereas our aesthetic field (green) produces smoother and more consistent scores that closely follow the teacher trend (blue). Importantly, this stability is \textit{not} achieved by explicit score smoothing, which would require arbitrary choices of window size or smoothness strength and lacks a principled basis without human ground-truth. Instead, our model \textit{implicitly} enforces smoothness through feature distillation, where learning in the aesthetic feature space naturally regularizes the score landscape across neighboring viewpoints.
These qualitative and quantitative results confirm that our distilled aesthetic field enables stable and consistent prediction of viewpoint-dependent aesthetics. More examples are provided in the Supplementary Material.
The success in novel view aesthetic prediction forms the foundation for aesthetic viewpoint suggestion, which we evaluate next.

\begin{figure*}[t]
\centering
    \includegraphics[width=\linewidth]{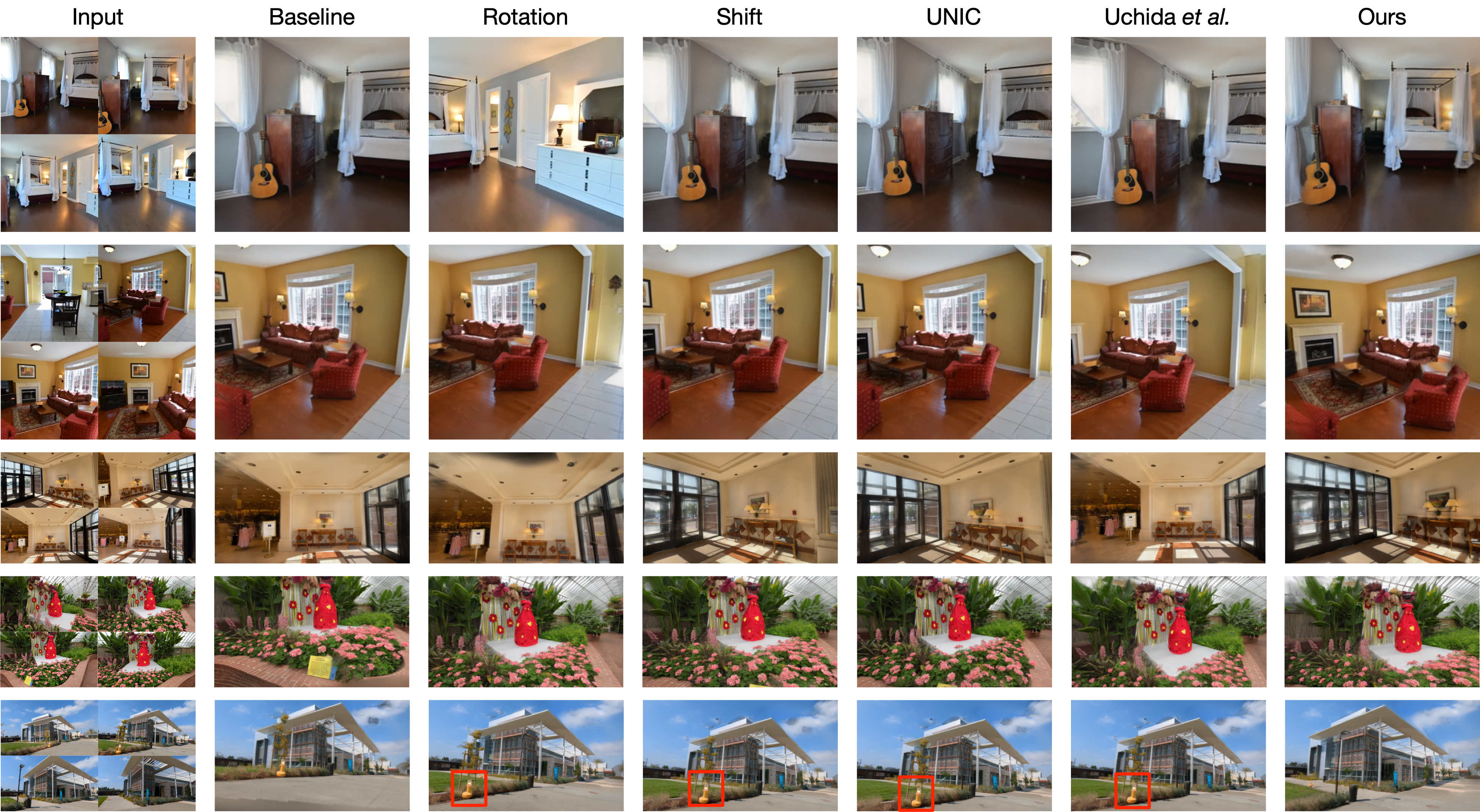}
    \caption{Aesthetic viewpoint suggestion examples with 4 input views in RE10k (\textit{top 2 rows}) and DL3DV (\textit{bottom 3 rows}). Red boxes in the last row show that single-view methods fail to remove distracting objects due to limited adjustment range. Zoom in for more details.}
    \label{fig:avs}
\end{figure*}

\subsection{Aesthetic Viewpoint Suggestion}
To evaluate our method on aesthetic viewpoint suggestion given sparse input views, we vary the number of input views between $[2,4,6]$ and assess the aesthetic quality of the suggested viewpoints. 

Since no existing benchmarks account for viewpoint-dependent aesthetics, we adopt the test splits of RE10k and DL3DV to provide input views and use two aesthetic models (VEN~\cite{wei2018good} and SAMPNet~\cite{sampnet}) to quantify framing and composition quality. Because the suggested views may not have corresponding ground-truth images, we reconstruct the scenes with dense inputs to obtain pseudo ground-truths for evaluation only, ensuring a fair and artifact-mitigated comparison without affecting the viewpoint suggestion process itself. As no prior work operates in this sparse-view setting, we compare against the baseline and single-view methods adapted to our evaluation protocol as detailed below.

\noindent\textbf{Comparison Approaches.} We first compare our method with the direct RGB-scoring baseline, which shares the same search pipeline as ours. To adapt single-view adjustment methods~\cite{li2025towards,su2021camera,uchida20253d,liu2023beyond,yao2025photography,sheng2024view} comparable in our setting, we treat each sparse input view as an anchor, around which we obtain a candidate based on their suggestions. This yields $N$ suggestions for the $N$ input anchors per scene, where we take the best as the final output. 
Among single-view methods, only UNIC~\cite{liu2023beyond} and Uchida \etal~\cite{uchida20253d} are publicly available. 
For fair comparison, we adapt both by fixing their crop sizes as input resolution and map their outpainted suggestions to corresponding real viewpoints. See Supplementary Material for additional details.
For other methods that predict minor \textit{in-plane shift}~\cite{su2021camera,sheng2024view} or \textit{rotation}~\cite{li2025towards} around the anchor viewpoint, we approximate their behavior by exhaustively sampling and scoring nearby views with the aesthetic model. Yao \etal\cite{yao2025photography} cannot be compared as their video generation model is not available. 3D-exploration methods are not comparable as they need dense captures of the scene.

\noindent\textbf{Quantitative Results.} \Cref{tab:avs} reports the aesthetic viewpoint suggestion results.  
Our method consistently recommends views with higher aesthetic scores than all comparison approaches in different metrics.
It also maintains superior performance across different input numbers, showing robustness under varying observation sparsity. 
Even with only two views, our model produces significantly better suggestions, showing its strong ability to reason about the underlying 3D scene aesthetics from minimal observations.
As the number of input views increases, the suggested viewpoints achieve progressively higher scores, indicating that our model effectively exploits broader scene coverage to discover more aesthetically optimal perspectives.
With more input views, the performance gap over single-view-based methods (\eg \textit{rotation}) narrows, likely because they can also explore sufficient viewpoint variations within their local neighborhoods. However, note that \textit{inplane-shift} and \textit{rotation} serve as an \textit{upper bound} for approximating single-view methods in our setting, as we directly maximize target scores rather than relying on learned guidance.

\noindent\textbf{Qualitative Comparisons.} \Cref{fig:avs} presents qualitative examples of the suggested viewpoints.  
Our framework identifies visually balanced and well-composed views that align with human aesthetic preferences. In contrast, single-view adjustment methods are restricted to small neighborhood around the anchor views and often fail to capture optimal perspectives. For example, they may break the structural continuity of objects (\eg bed frames in the first row) or struggle to remove distracting objects (\eg marked by red boxes in the last row) from the view while our method can freely place the camera to enhance composition.

\begin{figure}[t]
\centering
    \includegraphics[width=\linewidth]{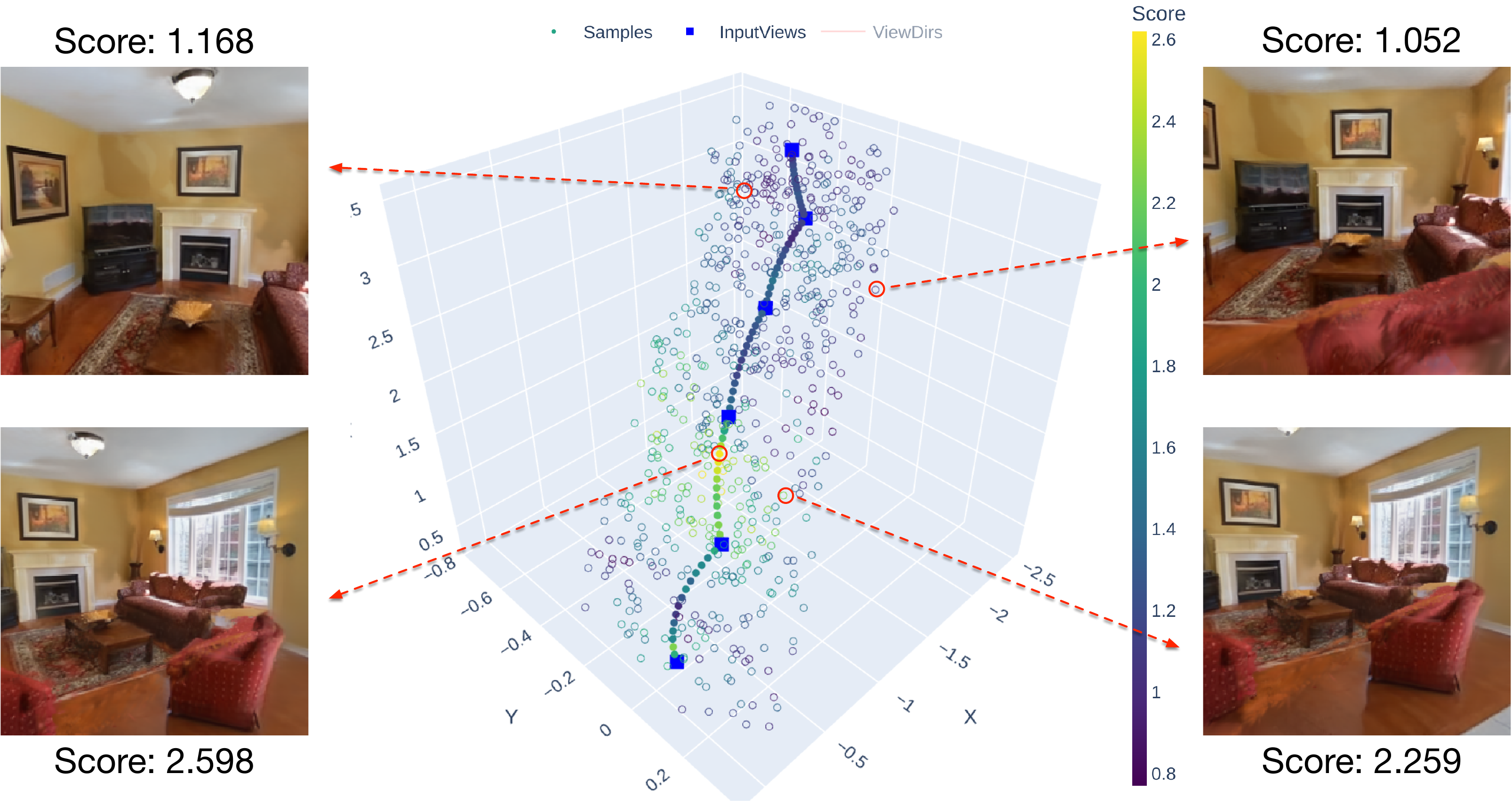}
    \caption{Visualization of sampled viewpoints colored by aesthetic score, with representative renderings shown alongside.}
    \label{fig:vis}
\end{figure}

\noindent\textbf{3D Visualization of Viewpoint Suggestions.} Finally, we visualize the viewpoint sampling strategy in~\cref{fig:vis}. Candidate viewpoints are sampled along and around the trajectory of input views, with colors indicating the predicted aesthetic scores. The results reveal how aesthetic quality varies across 3D space, and the corresponding renderings confirm strong alignment with human perceptual preferences.

\begin{table}[t]
\small
  \centering
  \begin{tabular}{l| c c}
    \toprule
    \multirow{2}{*}{Methods}  & RE10k~\cite{re10k} & DL3DV~\cite{dl3dv} \\
      & $\Delta$VEN$\uparrow$ & $\Delta$VEN$\uparrow$ \\
    \midrule
    Baseline & 0.20 & 0.18  \\
    \textbf{Ours} & \textbf{0.46} & \textbf{0.43} \\
    \bottomrule
  \end{tabular}
  \caption{Aesthetic score improvement with gradient ascent.}
  \label{tab:grad}
\end{table}

\subsection{Aesthetic Field Gradient Optimization Analysis}
\label{sec:grad}
\begin{figure}[t]
\centering
    \includegraphics[width=\linewidth]{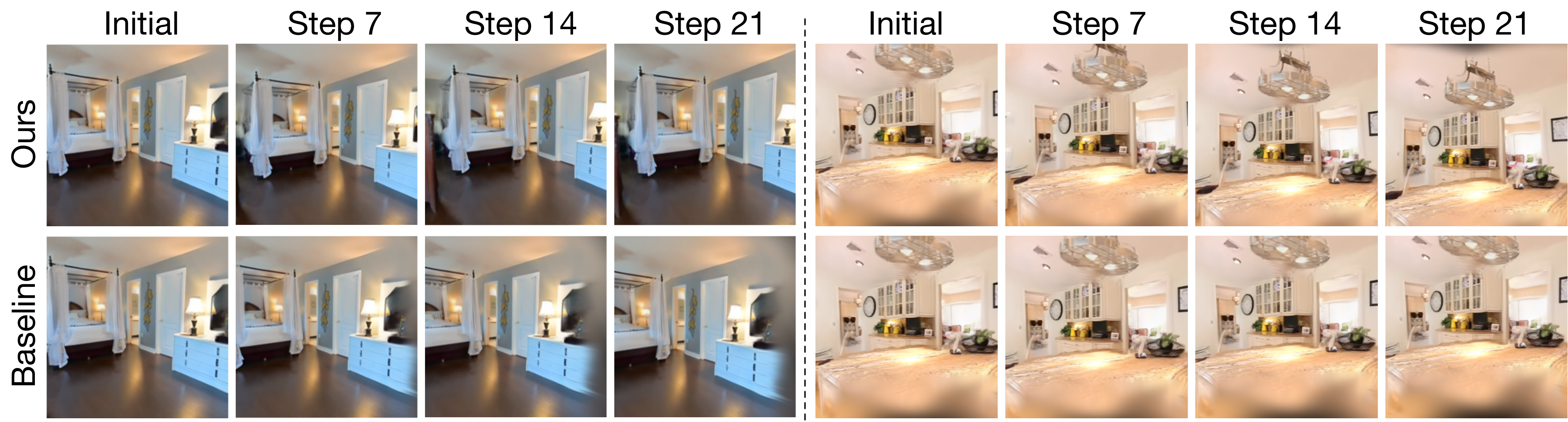}
    \caption{Gradient ascent results. Zoom in for a closer look.}
    \label{fig:grad}
\end{figure}

We analyze the learned aesthetic field through the lens of gradient-based optimization, and show that the distilled field enables stable and effective gradient ascent compared with the RGB-scoring baseline. We randomly select starting viewpoints and perform gradient ascent on both methods.
For each scene in the test split, both approaches receive the same sparse input views, and the same random novel view is chosen as the initial camera pose for both.  
We run optimization for $25$ steps and report the average score improvement $\Delta$VEN in \cref{tab:grad}.  
Our method achieves consistent score increases in different datasets, confirming that the learned aesthetic field provides a well-behaved optimization landscape.  
In contrast, the RGB-scoring baseline suffers from unstable updates that often degrade the results.
\Cref{fig:grad} further illustrate this behavior. Our method converges toward balanced and aesthetically pleasing viewpoints, while RGB-based refinement fails to update reliably toward better views. See Supplementary Material for more examples.

\subsection{Ablations}
We conduct ablation studies to validate key architectural and configuration choices in our framework.

\noindent\textbf{View-conditioning.} We first examine the role of view-conditioning in aesthetic prediction for novel viewpoints. \cref{tab:ablate_view} shows that view-conditioning significantly improves aesthetic prediction accuracy at unseen views. This confirms that explicitly modeling viewpoint dependence is crucial for capturing aesthetic cues across diverse perspectives.

\begin{table}[t]
\small
  \centering
  \begin{tabular}{l| c c | c c}
    \toprule
    \multirow{2}{*}{Methods}  & \multicolumn{2}{c|}{RE10k~\cite{re10k}} & \multicolumn{2}{c}{DL3DV~\cite{dl3dv}} \\
      & PLCC & SRCC & PLCC & SRCC \\
    \midrule
    w/o. view-cond. & 0.732 & 0.695 & 0.658 & 0.625  \\
    \textbf{w. view-cond.} & \textbf{0.796} & \textbf{0.758} & \textbf{0.700} & \textbf{0.668} \\
    \bottomrule
  \end{tabular}
  \caption{Ablation of view-conditioning on novel view aesthetic prediction with 4 input views.}
  \label{tab:ablate_view}
\end{table}

\noindent\textbf{Search Configurations.} We analyze the number $K$ of candidates returned in coarse sampling and gradient ascent steps in refinement. As shown in \cref{tab:ablate_search}, performance gains saturate around $K=2$ and 25 refinement steps, which are therefore adopted as default in all experiments. Additional ablations are provided in the Supplementary Material.

\begin{table}[t]
\small
  \centering
  \begin{tabular}{c | c | c}
    \toprule
    \multirow{2}{*}{\# Candidates} & RE10k~\cite{re10k} & DL3DV~\cite{dl3dv} \\
    & VEN$\uparrow$ & VEN$\uparrow$  \\
    \midrule
    1 & 1.96 & 2.57  \\
    2 &  2.03 & 2.76  \\
    3 &  2.05 & 2.78  \\
    \midrule
     \# Steps & $\Delta$VEN$\uparrow$ & $\Delta$VEN$\uparrow$ \\
    \midrule
    15 & 0.21 & 0.15 \\
    20 & 0.32 & 0.28 \\
    25 & 0.46 & 0.43 \\
    30 & 0.49 & 0.45 \\
    \bottomrule
  \end{tabular}
  \caption{Ablation of candidate numbers and refinement steps.}
  \label{tab:ablate_search}
\end{table}
\section{Discussion \& Conclusion}

We present a novel framework for 3D-aware aesthetic viewpoint suggestion that learns a 3D aesthetic field via feedforward 3D Gaussian Splatting, enabling geometry-grounded aesthetic reasoning and efficient aesthetic viewpoint discovery from sparse input views.
Despite its effectiveness, our framework still has room for improvement.
First, it relies on camera poses to build the aesthetic field. While such information can be obtained from COLMAP~\cite{schonberger2016structure} and is often available in robots/drones and smartphones~\cite{apple_arkit}, a pose-free variant would broaden the applicability of our method. This could be achieved by incorporating recent advances in pose-free methods~\cite{ye2024no,jiang2025anysplat}.
Second, the quality of the aesthetic field depends on the accuracy of the reconstructed geometry, which is affected by both the backbone’s ability and the input coverage. While the former can be improved by stronger geometry backbones~\cite{wang2025vggt}, the latter can be addressed by selecting views to ensure sufficient scene coverage~\cite{xiao2024nerf,chen2024gennbv}. 
Third, our viewpoint search is constrained to regions supported by the initial observations. A promising extension is an active-perception loop that acquires additional views in promising directions to expand the aesthetic field and enlarge the feasible search space.

\clearpage
\setcounter{page}{1}
\setcounter{figure}{0}
\setcounter{table}{0}
\renewcommand{\thefigure}{S\arabic{figure}}
\renewcommand{\thetable}{S\arabic{table}}
\maketitlesupplementary

\setcounter{section}{0} 
\renewcommand{\thesection}{S\arabic{section}} 
\setcounter{subsection}{0} 
\renewcommand{\thesubsection}{S\arabic{section}.\arabic{subsection}} 

\section{Datasets \& Implementation Details}
\label{sec:sup_impl}
We conduct experiments on subsets of the RE10k~\cite{re10k} and DL3DV~\cite{dl3dv} datasets. For RE10k, we use approximately $6,000$ scene clips as the training split and $749$ clips as the test split. For DL3DV, we adopt its 1K-5K subsets for training, which contain roughly $5,000$ scenes, and test on the $140$ benchmark scenes.

During training, we sample 2 input views to reconstruct the aesthetic features of 12 target views in RE10k, and sample 2 to 6 input views to predict aesthetic features for 24 target views in DL3DV. We train all models for 50,000 steps with a batch size of 8. We use AdamW~\cite{loshchilov2017decoupled} with a learning rate of $0.0005$ and a cosine annealing scheduler.

\begin{figure}[h]
\centering
\begin{subfigure}[t]{0.98\linewidth}
\captionsetup{labelformat=empty}
    \centering
    \includegraphics[width=\linewidth]{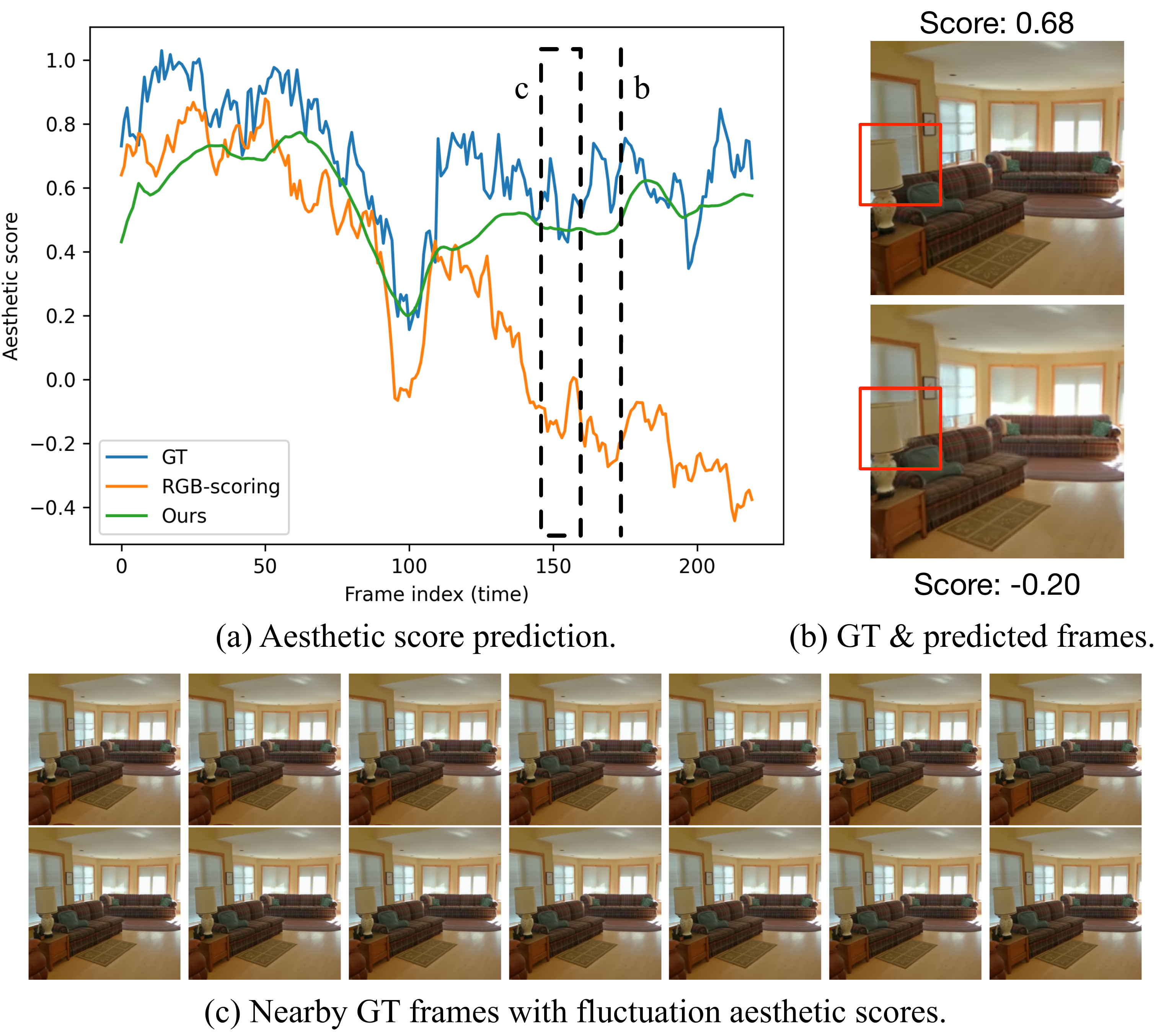}
    \caption{
        \textbf{Sequence 1.}
    }
\end{subfigure}
\vfill
\begin{subfigure}[t]{0.98\linewidth}
\captionsetup{labelformat=empty}
    \centering
    \includegraphics[width=\linewidth]{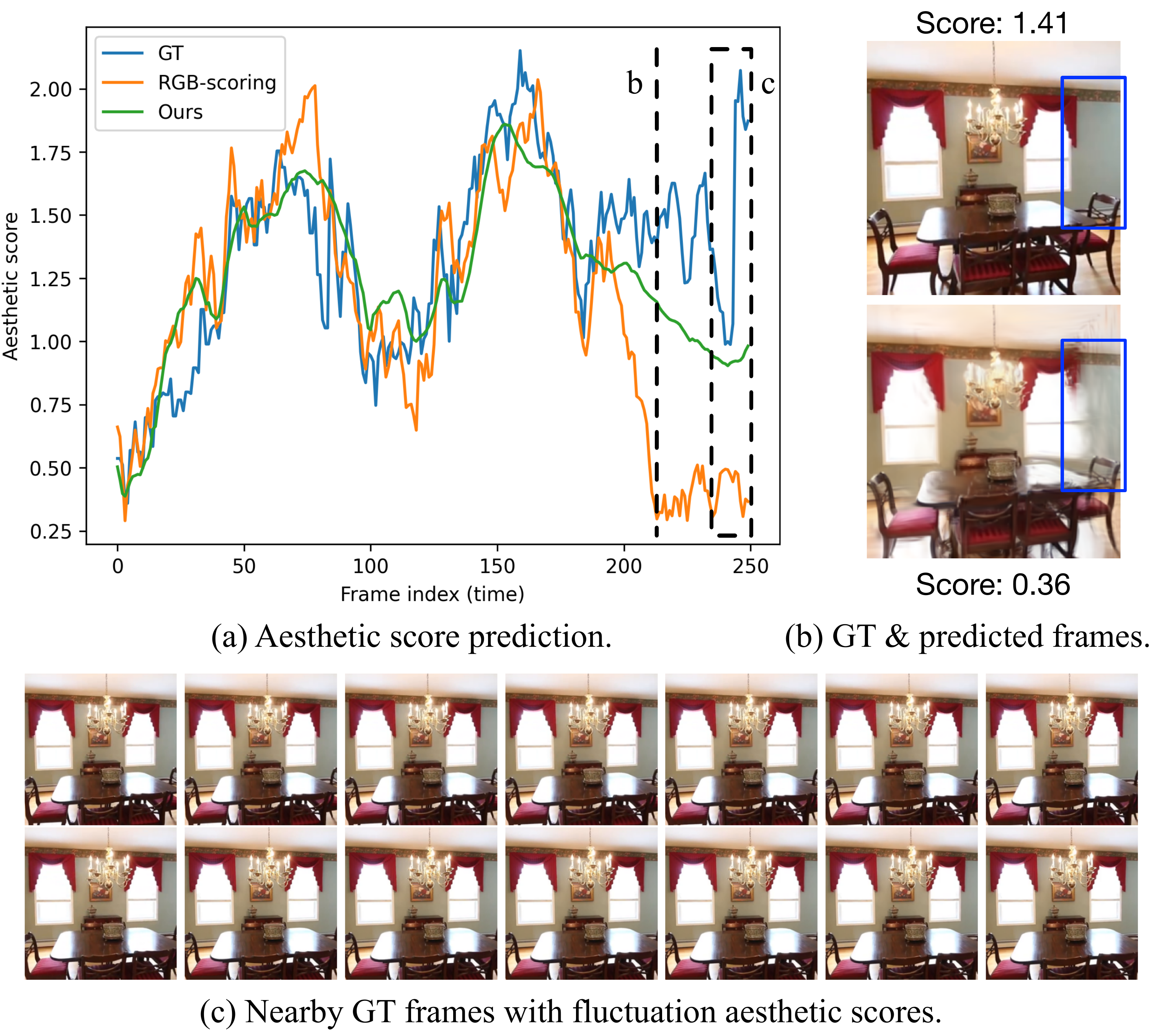}
    \caption{
        \textbf{Sequence 2.}
    }
\end{subfigure}

\caption{
Additional novel view aesthetic prediction results. \textit{In both examples:} (a) Aesthetic score predictions over consecutive frames. The dashed line and box mark the regions visualized in (b) and (c), respectively.
(b) Given the same aesthetic model and viewpoint, rendering artifacts in the predicted view (\eg, wrong color prediction as marked in red boxes in \textit{top}, noisy predictions marked in blue boxes in \textit{bottom}) bias the RGB-scoring approach.  
(c) Ground-truth scores fluctuate noticeably across nearly identical nearby views. 
}
\label{fig:sup_nva}
\end{figure}

\section{Aesthetic Prediction at Novel Views}
\label{sec:sup_nva}
We provide additional qualitative examples of novel-view aesthetic prediction to further illustrate the two limitations of the RGB-based baseline approach discussed in the main paper. As shown in~\cref{fig:sup_nva}, the aesthetic model is biased by the rendering artifact, and exhibits high sensitivity to minor pixel-level variations across nearby views that are nearly identical. In contrast, our distilled aesthetic field faithfully follows the teacher's underlying trend, producing smoother and more consistent predictions across nearby frames.

\begin{figure*}[htbp]
\centering
    \includegraphics[width=\linewidth]{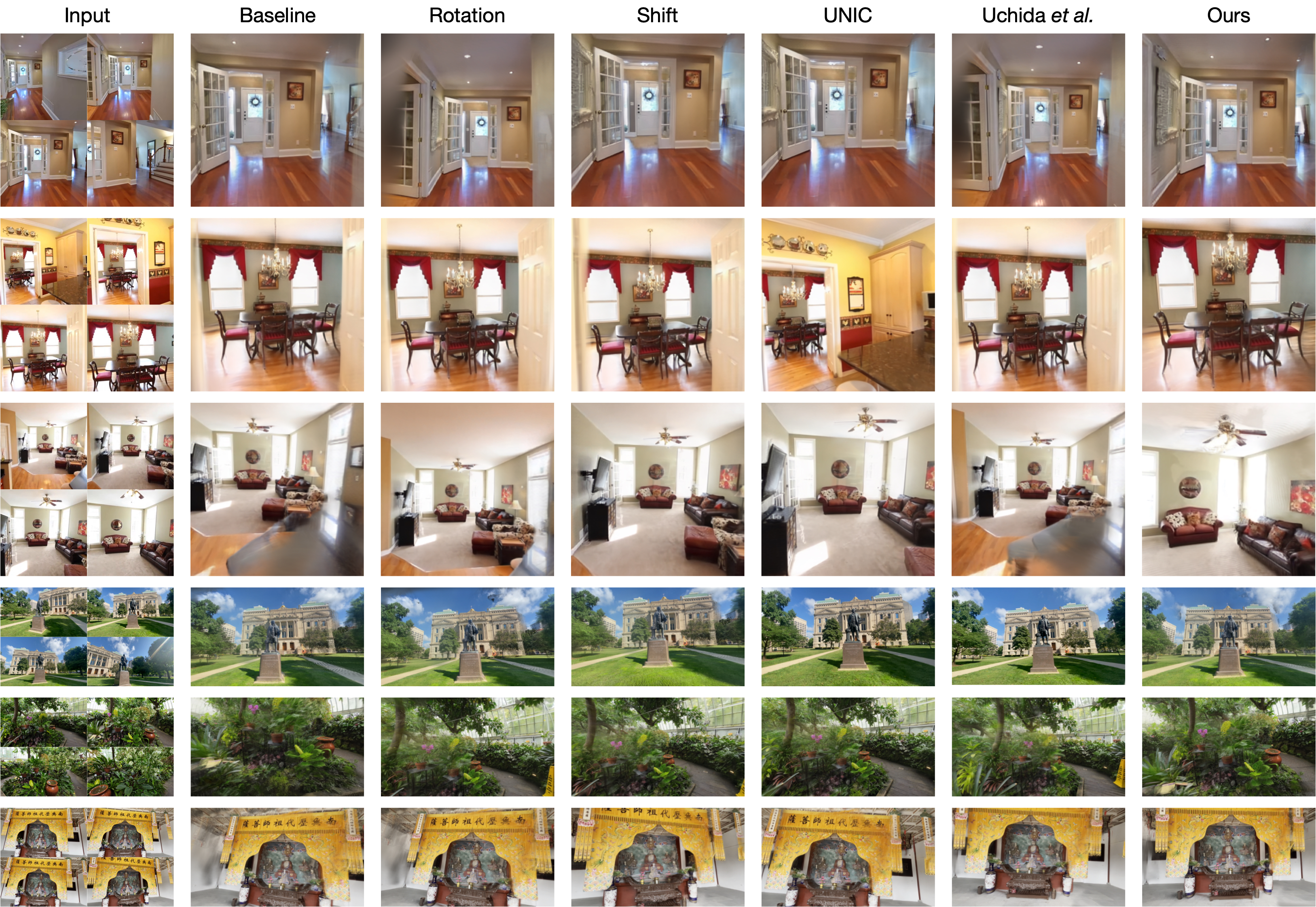}
    \caption{Additional results of aesthetic viewpoint suggestion.}
    \label{fig:sup_avs}
\end{figure*}

\begin{figure*}[p]
\centering
    \includegraphics[width=\linewidth]{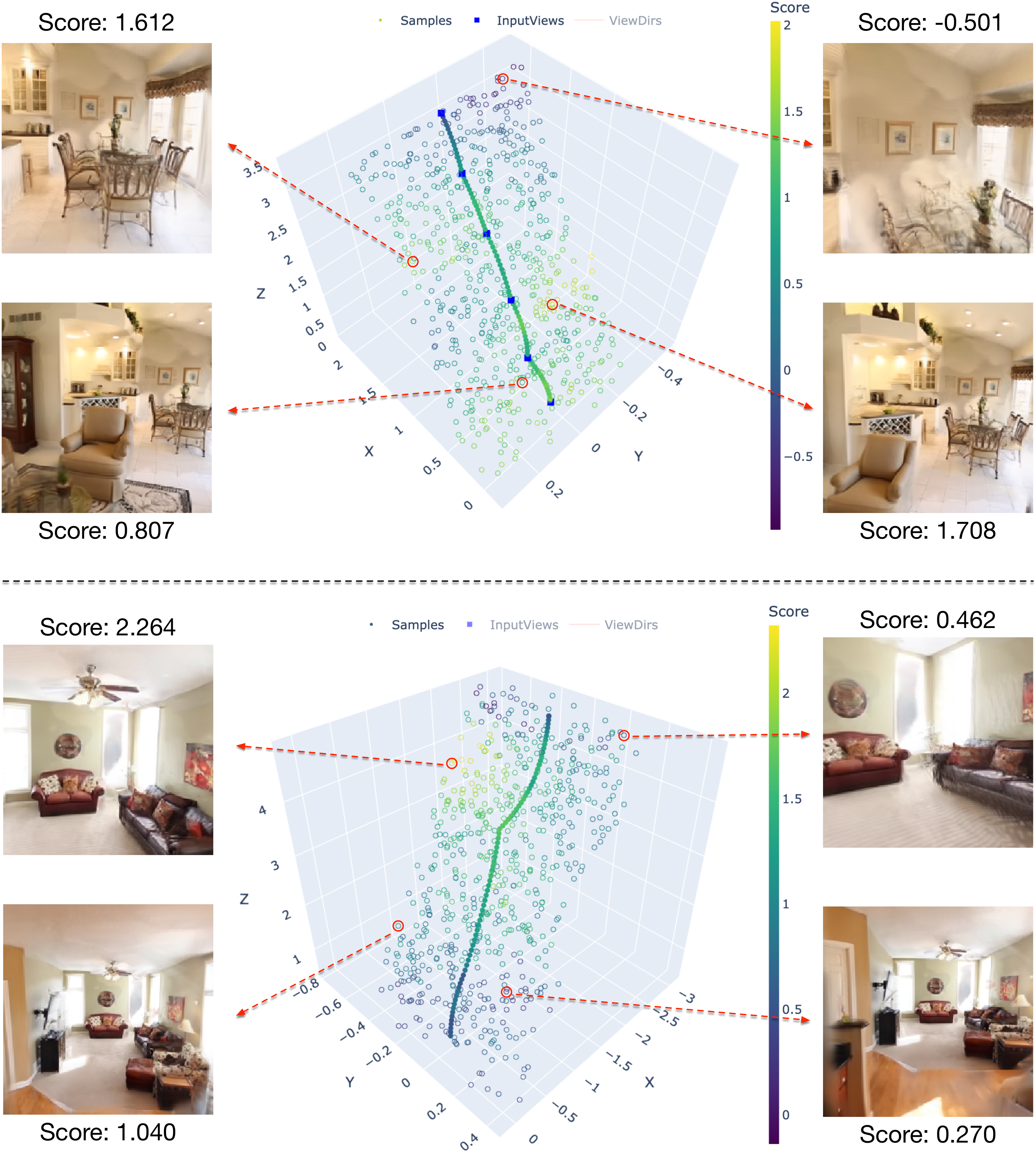}
    \caption{Additional visualizations of sampled viewpoints colored by aesthetic score, with representative renderings shown alongside.}
    \label{fig:sup_vis}
\end{figure*}

\section{Aesthetic Viewpoint Suggestion}
\label{sec:sup_avs}
We first describe how existing single-view baselines are adapted to our evaluation setting, followed by additional qualitative comparisons of the suggested viewpoints.

UNIC~\cite{xu2023unifying} predicts crops with varying sizes in an extrapolated image plane rather than real camera motions. For fair comparison, we fix its crop size as input resolution ($256\times 256$ in RE10k and $256\times 448$ in DL3DV) and map the crop-center shifts to in-plane camera translations, thereby obtaining corresponding real viewpoints where we measure aesthetic qualities. Uchida \etal~\cite{uchida20253d} first outpaint an image, reconstruct 3D point clouds from it, and then optimize camera poses and image aspect ratios in this outpainted 3D scene. We therefore also fix their output resolution and transform their suggestions to equivalent real viewpoints for evaluation. 

As shown in~\cref{fig:sup_avs}, our method consistently suggests viewpoints with superior framing and composition than comparison methods.

Finally, additional 3D visualizations of the viewpoint sampling results are shown in~\cref{fig:sup_vis}. The results demonstrate the variation of aesthetic quality across 3D space, and the corresponding renderings confirm strong alignment between our model's evaluations and human perceptual preferences.

\begin{figure*}[t]
\centering
    \includegraphics[width=0.92\linewidth]{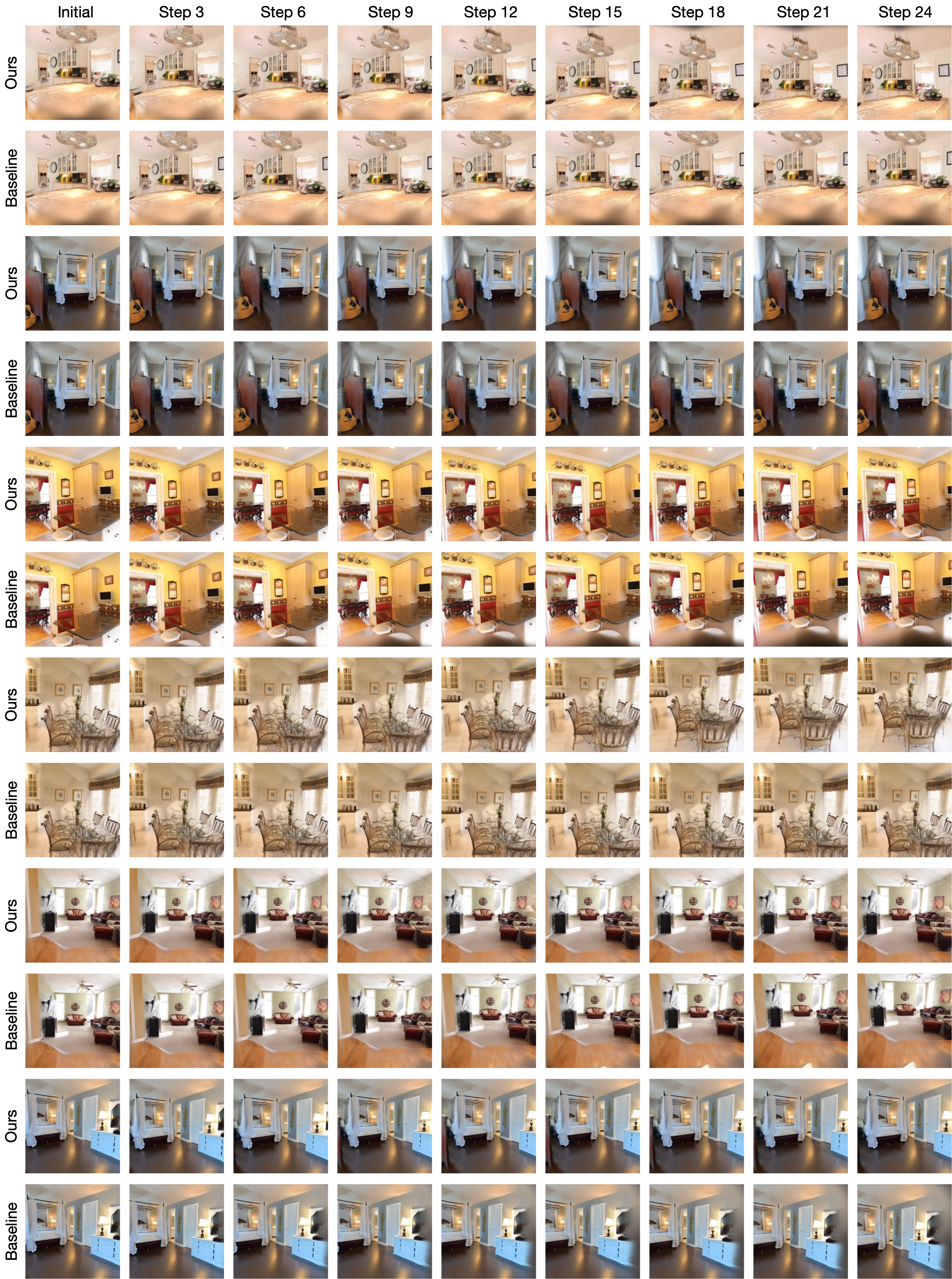}
    \caption{Additional gradient ascent results. Zoom in for a closer look.}
    \label{fig:sup_grad}
\end{figure*}

\section{Aesthetic Field Gradient Optimization}
We present additional results comparing gradient ascent optimization using the baseline approach and our method. As shown in~\cref{fig:sup_grad}, our method consistently converges toward more aesthetically pleasing viewpoints, whereas the baseline approach often fails to make meaningful progress and can sometimes degenerate.

\begin{table}[t]
\small
  \centering
  \begin{tabular}{l| c c  c c  c}
    \toprule
    & $S=4$& $S=8$& $S=16$& $S=32$\\
    \midrule
    $N=4$ & 1.98 & 2.00 & 1.98 & 1.98  \\
    $N=8$ & 1.98 & 1.94 & 2.03 & 2.02 \\
    $N=16$ & 2.01 & 2.00 & 2.02 & 2.04  \\
    $N=32$ & 2.02 & 2.03 & 2.04 & 2.06  \\
    \bottomrule
  \end{tabular}
  \caption{Ablation of the sampling configuration of search stage 1 on novel view aesthetic prediction. $S$ denotes the number of samples along each segment of the interpolated input-view trajectory, and $N$ denotes the number of neighboring viewpoints sampled around each segment-sample. All experiments use 4 input views on RE10k~\cite{re10k}, and the aesthetic quality is measured by VEN~\cite{wei2018good}.}
  \label{tab:sup_ablate}
\end{table}

\section{Ablations}
We provide additional ablation experiments of the Stage 1 search configuration, focusing on two parameters: the number of samples along each input-view segment ($S$) and the number of neighbors further drawn around each segment sample ($N$). The total number of candidate viewpoints is approximately proportional to $S\times N$. We measure the aesthetic qualities of final suggestions via VEN~\cite{wei2018good}. As shown in ~\cref{tab:sup_ablate}, our method remains stable across a wide range of sampling densities, while generally achieving better performance with more samples. We adopt $S=16$ and $N=8$ as our default configuration, as this setting offers a good balance between suggestion quality and total number of samples being evaluated.

%
{
    \small
    \bibliographystyle{ieeenat_fullname}
    \bibliography{main}
}
\end{document}